\documentclass[10pt,twocolumn,letterpaper]{article}

\usepackage{cvpr}              

\usepackage{caption}
\usepackage{graphicx}
\usepackage{amsmath}
\usepackage{amssymb}
\usepackage{booktabs}
\usepackage{enumitem}
\usepackage{cuted}
\usepackage{multirow}
\usepackage{xcolor}

\definecolor{cvprblue}{rgb}{0.21,0.49,0.74}
\usepackage[pagebackref,breaklinks,colorlinks,allcolors=cvprblue]{hyperref}

\def\paperID{*****} 
\def\confName{CVPR}
\def\confYear{2026}

\title{CMAG: Concept-Scaffolded Retrieval for Marketplace Avatar Generation}

\author{
Rajeev Goel$^{1,2}$\thanks{Corresponding author. Work done during an internship at Roblox Corporation.} \quad 
Jason Ding$^{2}$ \quad 
Phani Harish Wajjala$^{2}$ \quad 
Pavan Turaga$^{1}$ \\
Tejaswi Gowda$^{1}$ \quad 
Krishna C. Garikipati$^{2}$ \\
$^{1}$Arizona State University \quad $^{2}$Roblox Corporation \\
}

\begin{document}
\maketitle

\begin{figure*}[t]
    \centering
    \includegraphics[width=\textwidth]{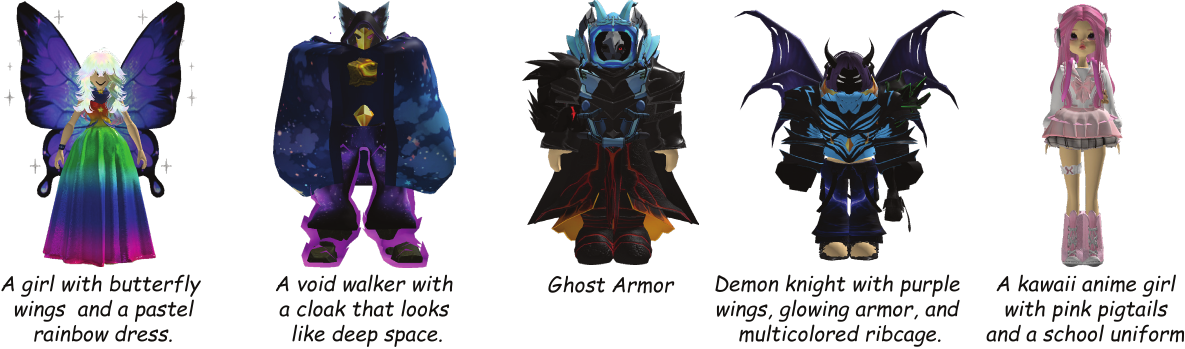}
    \caption{Illustration of high-quality avatars generated by our method from text prompts. Our approach enables versatile and fast generation, while accurately capturing fine-grained details from the prompts.}
    \label{fig:pipeline}
\end{figure*}

\begin{abstract}
Metaverse platforms rely on creator-driven marketplaces where avatars are assembled from discrete, taxonomy-labeled 3D assets (e.g., tops, bottoms, shoes, accessories) under strict category and topology constraints. While users increasingly expect free-form text control, text-only retrieval is brittle: natural language is ambiguous with respect to platform taxonomies, metadata is often noisy or informal, and independently retrieved components can be stylistically inconsistent or geometrically incompatible. We propose \textbf{CMAG}, a concept-scaffolded retrieval and verified composition framework for marketplace avatar generation. Given a prompt, CMAG first synthesizes an intermediate 3D concept scaffold that disambiguates intent beyond text by providing global spatial and stylistic context. In parallel, a view-aware part discovery module extracts localized visual evidence via prompt decomposition and text-grounded segmentation. A prompt-conditioned taxonomy router enforces category coverage and resolves semantic-to-taxonomic mismatch, after which a hybrid category-wise retriever combines part-based fusion with a concept-residual fallback using feature suppression. Finally, an agentic vision--language model filters and re-ranks candidates across categories and drives an iterative verification loop to assemble prompt-faithful, topologically consistent avatars from catalog assets.  We evaluate CMAG on diverse compositional prompts and demonstrate improved retrieval robustness and compositional correctness compared to strong baselines, highlighting the importance of 3D concept scaffolding under prompt ambiguity.
\end{abstract}    
\section{Introduction}
\label{sec:intro}

As social virtual worlds and metaverse platforms mature, \emph{3D avatars} have become the primary interface for identity, expression, and interaction. Modern platforms increasingly operate as creator-driven economies: users design and publish shirts, pants, accessories, and other avatar items to marketplaces where others can purchase and trade them. This ecosystem thrives when high-quality user-generated content (UGC) remains valuable and discoverable, yet users also expect avatar creation to be as effortless as describing what they want (e.g., ``a cyberpunk streetwear outfit with a black hoodie, cargo pants, and tactical headphones''). Bridging this gap is challenging: fully generative text-to-3D tools can bootstrap identity quickly, but they risk bypassing the marketplace, weakening creator incentives, and producing outputs that are not directly executable under platform-specific asset taxonomies and category-based avatar rigs.

A natural alternative is retrieval-based avatar generation, where the system composes an avatar by selecting compatible assets from an existing catalog. This paradigm preserves marketplace economics and guarantees executability, since outputs are constrained to real, platform-compliant assets. However, text alone is an unreliable bridge between user intent and discrete catalog items. Marketplace metadata is often sparse, informal, multilingual, or stylistically idiosyncratic, while user prompts are frequently ambiguous. More fundamentally, the mapping from free-form language to platform taxonomies is not one-to-one: a concept such as ``hoodie'' may correspond to multiple valid categories (e.g., ``sweater'' or ``jacket''), and different categories can share overlapping textual descriptions. Consequently, text-only embedding retrieval often returns plausible but taxonomically incorrect items, omits required components, and produces stylistically inconsistent combinations that break cross-category coherence.

This raises a central question: \emph{can foundation models improve retrieval robustness while still preserving creator-driven marketplaces?} Inspired by traditional 3D workflows in which artists first establish a global concept and then refine individual components, we argue that visual context is the missing signal for ambiguity-resilient retrieval. Rather than relying solely on text, we first instantiate a coarse 3D concept scaffold from the prompt. This scaffold is not the final output; instead, it provides a global visual prior over silhouette, garment layout, and accessory placement that disambiguates intent and guides downstream retrieval. Nevertheless, retrieving discrete assets from a holistic concept remains non-trivial: different components are best observed from different viewpoints, some parts may be occluded, and independently retrieved categories may conflict aesthetically or topologically. Marketplace avatar generation must therefore jointly address (i) \emph{semantic-to-taxonomic alignment}, (ii) \emph{robust evidence extraction} under occlusion and noisy metadata, and (iii) \emph{cross-category compatibility} under platform topology constraints.

To this end, we propose CMAG, a multi-stage agentic framework for concept-scaffolded retrieval and verified composition of marketplace avatars from free-form text. Given a prompt, CMAG first generates an intermediate 3D concept scaffold and renders multiple orthogonal views to obtain global visual embeddings. In parallel, a view-aware part discovery module decomposes the prompt into segmentable units, selects informative viewpoints per category, and extracts localized part embeddings via text-grounded segmentation when possible. A prompt-conditioned taxonomy routing policy resolves ambiguity and enforces category coverage by producing a target category set and stabilized per-category queries. Retrieval is then performed per category using a hybrid engine that (i) fuses localized part evidence with textual priors when available and (ii) falls back to concept-conditioned global retrieval via feature suppression when part evidence is unreliable. Finally, an agentic vision--language model enforces categorical correctness and cross-category compatibility through filtering and reranking, and an iterative verification loop assembles executable avatars while correcting missing components or geometric inconsistencies.

Our contributions can be summarized as:
\begin{itemize}
    \item We introduce a ``concept-scaffolded retrieval paradigm'' for marketplace avatar generation that transforms text-only asset search into a 3D-aware compositional process. By first instantiating an intermediate 3D concept scaffold, our framework establishes global spatial and stylistic structure before resolving discrete asset selection, mitigating semantic-to-taxonomic ambiguity inherent in free-form prompts.
    \item We propose a structured multi-stage framework that integrates view-aware part discovery, prompt-conditioned taxonomy routing, and a hybrid retrieval mechanism with low-rank feature suppression. This design isolates category-specific semantic subspaces, preserves coverage under occlusion or noisy metadata, and enforces cross-category topological compatibility during avatar assembly via an agentic vision--language verification loop.
    \item We demonstrate through extensive quantitative and qualitative evaluation on complex compositional prompts that CMAG achieves superior prompt faithfulness, structural coherence, and retrieval robustness compared to state-of-the-art avatar generation and semantic retrieval baselines.
\end{itemize}
\section{Related Works}
\label{sec:related_works}

\subsection{3D Generation and Agentic Frameworks}

Recent advances in text-to-3D generation have been driven by lifting powerful 2D diffusion priors into 3D representations~\cite{poole2022dreamfusion, nath2024deep, nath2025decompdreamer}, followed by efficient feed-forward architectures that enable rapid 3D synthesis~\cite{xiang2025trellis, hunyuan3d2025}. These approaches demonstrate impressive visual fidelity and scalability, showing that high-quality geometry and appearance can be generated directly from natural language prompts. However, fully generative pipelines typically produce unconstrained meshes or neural fields, which may not adhere to platform-specific topology, rig compatibility, or discrete asset taxonomies required in large-scale creator-driven marketplaces. In such User-Generated Content (UGC) ecosystems, avatars are composed from pre-authored assets that must satisfy strict stylistic and structural constraints. Retrieval-based pipelines naturally ensure executability by selecting existing catalog items, but they suffer from a substantial semantic gap between free-form language and discrete metadata. Ambiguous phrasing, informal descriptions, and non-unique mappings between concepts and categories often lead to noisy or incomplete retrieval results. This tension between generative flexibility and marketplace compliance motivates hybrid paradigms that combine generative priors with retrieval mechanisms to improve structural validity while preserving asset constraints.

Concurrently, the growing complexity of compositional 3D reasoning has led to the emergence of multi-agent architectures powered by Large Language Models (LLMs) and Vision-Language Models (VLMs). Frameworks such as LL3M~\cite{lu2025ll3m} decompose 3D generation into planning, retrieval, and execution agents, while CADCodeVerify~\cite{alrashedy2024generating} employs VLM-based verification loops for iterative geometric correction. More recently, AVATAR-AGENT~\cite{ding2026avataragent} demonstrates hierarchical planning and VLM-driven refinement for assembling avatars from discrete creator catalogs. These systems highlight the benefits of structured decomposition and iterative verification in complex 3D pipelines. 
In contrast to purely generative or purely retrieval-based approaches, our work integrates a 3D concept scaffold with taxonomy-aware retrieval and agentic verification, explicitly targeting marketplace-constrained avatar assembly under prompt ambiguity.

\subsection{Low-rank Learning}
Low-rank representations provide a principled framework for identifying compact and semantically meaningful subspaces within high-dimensional latent embeddings. Prior work has shown that deep neural networks often encode salient semantic structure in low-rank feature components, which can improve robustness and signal-to-noise separation in downstream tasks \cite{wang2024learning}. In generative settings, isolating such subspaces enables targeted concept manipulation and erasure without substantially degrading unrelated content. Recent methods exploit this property for controlled semantic editing. For example, GLoCE~\cite{lee2025gloce} employs gated low-rank adaptation to suppress localized concepts while preserving surrounding structure. Similarly, IP-Composer~\cite{dorfman2025ip} demonstrates that compositional reasoning can be performed directly in latent space by projecting embeddings onto text-identified CLIP subspaces and combining orthogonal components. Motivated by these observations, our hybrid retrieval engine adopts a low-rank subspace projection strategy to suppress non-target semantic directions in global view embeddings. By orthogonalizing category-specific residuals against subspaces corresponding to other asset categories, we reduce interference from adversarial multi-concept noise and improve retrieval robustness under ambiguous prompts.
\section{Method}
\label{sec:method}

\begin{figure*}
    \centering
    \includegraphics[width=\textwidth]{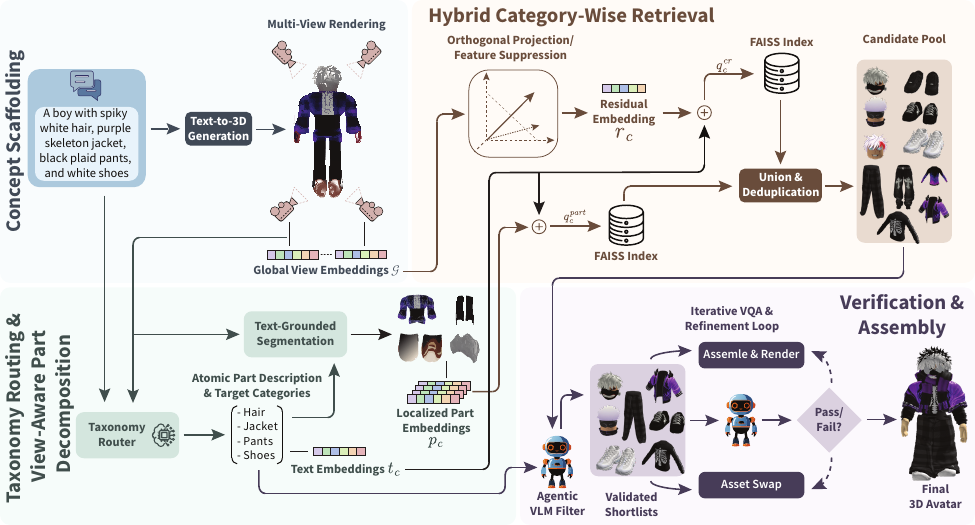}
    \caption{Overview of the CMAG pipeline. Given a user prompt, we construct a 3D concept scaffold and perform view-aware part decomposition to extract global and localized visual evidence. A prompt-conditioned taxonomy router determines target categories and stabilized retrieval queries. We then perform hybrid category-wise retrieval via part-based querying and global concept erasure. Retrieved candidates are filtered and combinatorially reranked by an agentic VLM before iterative avatar assembly, verification, and final selection.}
    \label{fig:pipeline}
    \vspace{-0.5em}
\end{figure*}

\subsection{Overview}
\label{sec:method_overview}
Our aim is to synthesize visually coherent, topologically consistent 3D avatars from free-form natural language prompts through the retrieval and composition of discrete catalog assets, enforcing strict taxonomy alignment and prompt faithfulness. The proposed modular pipeline is illustrated in Figure~\ref{fig:pipeline}. A fundamental challenge in this domain is the semantic-to-taxonomic misalignment. Free-from user prompts frequently maps ambiguously to multiple valid discrete categories (e.g., a ``hoodie'' may correspond to either \texttt{sweater} or \texttt{jacket}). Furthermore, upstream segmentation processes are susceptible to omission or mislabeling, and downstream zero-shot embedding retrieval, being inherently recall-oriented, often retrieves plausible yet stylistically or semantically incompatible assets. To systematically mitigate these failure modes, we introduce a coarse-to-fine compositional generation framework comprising five tightly coupled modules: (i) a prompt-conditioned 3D concept scaffold establishing global visual context; (ii) view-aware part discovery for extracting localized visual evidence; (iii) a prompt-conditioned taxonomy routing policy to ensure category coverage and semantic disambiguation; (iv) a hybrid retrieval mechanism supporting both part-based querying and orthogonalized concept-erasure; and (v) a Vision-Language Model (VLM)-based critique and verifier to enforce rigorous categorical and aesthetic correctness for the final avatar generation.

\subsection{Concept Scaffolding via Prompt-to-3D and Multi-View Encoding}
\label{sec:concept_scaffold}
\subsubsection{Prompt-to-3D concept generation.}
Given an input text prompt, we synthesize a textured 3D mesh that functions as an \emph{intermediate concept scaffold} rather than the final topological output. The primary utility of this scaffold is to anchor global appearance cues; such as silhouette, coarse garment geometry, and accessory placement, and to facilitate multi-view rendering for subsequent view-aware perception.

\subsubsection{Multi-view rendering and global visual embeddings.}
From the generated 3D concept scaffold, we render four orthogonal views and independently encode each using a CLIP-style image encoder. We intentionally eschew feature pooling or averaging operations across views, as asymmetric feature visibility dictates that different asset categories are optimally represented from distinct viewpoints (e.g., back accessories are maximally visible from the posterior view, whereas facial accessories necessitate front views). We formalize the resulting set of global view embeddings as
\begin{equation}
\mathcal{G} = \{\mathbf{g}_v\}_{v \in \mathcal{V}},
\end{equation}
where $\mathcal{V}$ denotes the set of rendered viewpoints. These embeddings furnish category-conditional priors when detailed visual evidence is unavailable, and they establish the global spatial context for subsequent view-aware part discovery.

\subsection{Ambiguity Resolution via Taxonomic Routing}
\label{sec:taxonomy_router}
A prominent source of retrieval failure is \emph{taxonomic mismatch}: a prompted concept may map to multiple valid catalog categories, and downstream part discovery may omit a requisite category entirely. To systematically resolve this, we introduce a prompt-conditioned taxonomy routing policy, executed prior to retrieval, which explicitly optimizes for hierarchical recall and topological coherence.

Given the prompt, the catalog taxonomy, and the prior decomposition outputs, the router generates two distinct artifacts:
\begin{enumerate}
    \item \textbf{A target category set} $\mathcal{C}_p$ delineating the specific catalog categories that must be retrieved to satisfy the prompt and ensure visual completeness.
    \item \textbf{A retrieval query} $q_c$ for each selected category $c \in \mathcal{C}_p$, engineered for maximum contrastive alignment in the CLIP latent space by isolating the key noun and appending discriminative modifiers (e.g., material, color, and aesthetic theme).
\end{enumerate}

The routing policy operates on two fundamental principles. First, it enforces a \emph{recall-preserving} strategy under ambiguity: if a concept plausibly maps to multiple categories, all such categories are included to prevent the premature exclusion of valid assets. Second, it enforces \emph{taxonomic coherence}: when prompts dictate mutually exclusive topological decisions (e.g., \texttt{jacket} versus \texttt{sweater}), the router deterministically selects the consistent category to prevent downstream geometric intersections. The output queries $q_c$ serve as the stabilized text conditioning signals for the retrieval engine and the grounded segmentation.

\subsection{View-Aware Part Decomposition and Grounded Segmentation}
\label{sec:part_decomposition}
Avatar descriptions in natural language are inherently compositional, often specifying a conjunction of atomic entities (e.g., ``tactical headphones'', ``cargo pants''). To map this continuous textual intent to localized visual evidence, we employ a view-aware part decomposition module consisting of three integrated components.

\subsubsection{Prompt decomposition into segmentable units.}
Utilizing a pre-trained Vision-Language Model (VLM), the input prompt is decomposed into a discrete set of atomic part descriptions, conditioned on the target avatar taxonomy obtained from the taxonomy agent(Sec.~\ref{sec:taxonomy_router}). The generated phrases are constrained to be both visually grounded (i.e., spatially segmentable) and structurally aligned with valid taxonomy concepts.

\subsubsection{Category-conditioned view selection.}
Due to self-occlusion and the viewpoint-dependency of part visibility, the decomposition module dictates the optimal observation plane(s) for each category. For instance, back accessories are preferentially segmented from the posterior view, while waist accessories may necessitate lateral views to mitigate frontal silhouette ambiguity. This category-conditioned routing significantly curtails false negatives caused by uninformative camera angles.

\subsubsection{Text-grounded segmentation and part embeddings.}
We apply open-vocabulary text-grounded segmentation on the optimal view(s) to isolate category-specific crops. To enhance robustness, failure to generate a usable mask triggers a fallback mechanism that augments the phrase with broader taxonomy keywords. Successful crops are subsequently passed through the CLIP-style image encoder to extract localized part-level embeddings $\{\mathbf{p}_c\}$. Despite these mechanisms, segmentation may still fail for highly occluded or abstract items; consequently, we rely on the subsequent taxonomy routing and hybrid retrieval stages to guarantee semantic coverage.

\subsection{Hybrid Category-Wise Retrieval and Latent Concept Erasure}
\label{sec:retrieval}
We design a hybrid retrieval engine to assemble a high-recall, precision-bounded pool of candidate assets per routed category, maintaining robustness against missing or occluded part crops.

\subsubsection{Notation and category-wise indices}
Let $\mathcal{C}$ define the universe of asset categories, and let $\mathcal{D}_c$ denote the subset of catalog items belonging to category $c \in \mathcal{C}$. We precompute $L_2$-normalized CLIP-style image embeddings for the entire catalog and construct independent FAISS indices per category, restricting nearest-neighbor search to the taxonomy-consistent candidate set.

\subsubsection{Localized Part-Based Retrieval}
When segmentation yields a valid crop for category $c$, we extract its part embedding $\mathbf{p}_c$. The router-provided query text $q_c$ is encoded to obtain the textual prior $\mathbf{t}_c$. We fuse the localized visual evidence with the textual intent via a weighted linear combination:
\begin{equation}
\mathbf{q}^{\text{part}}_c = \alpha\,\mathbf{p}_c + (1-\alpha)\,\mathbf{t}_c,
\end{equation}
where $\alpha$ modulates the reliance on visual versus textual modalities. We query the category-specific index with $\mathbf{q}^{\text{part}}_c$ to obtain the precision-oriented candidate set $\mathcal{R}^{\text{part}}_c$.

\subsubsection{Global Retrieval via Feature Suppression}
To preserve retrieval coverage when part discovery fails, we derive a category-targeted query directly from the global view features by orthogonalizing against non-target categories. Let $\mathbf{g}$ denote the selected global view embedding for target category $c$, and $d$ be the embedding dimension.

We precompute a low-rank projection matrix $\mathbf{P}_k \in \mathbb{R}^{d \times d}$ for each category $k$ using Singular Value Decomposition (SVD) over the catalog embedding covariance matrix, retaining the top-$r$ principal directions to define the category subspace. During inference, we derive a category-conditioned residual embedding $\mathbf{r}_c$ by iteratively projecting $\mathbf{g}$ onto the orthogonal complement of all other categories:
\begin{equation}
\mathbf{r}_c \leftarrow \mathbf{g}, \qquad
\mathbf{r}_c \leftarrow \mathbf{r}_c - \mathbf{P}_k\,\mathbf{r}_c \;\;\; \forall k \in \mathcal{C},\; k \neq c.
\end{equation}
This orthogonalization actively suppresses adversarial multi-concept noise. We fuse this residual with the textual prior:
\begin{equation}
\mathbf{q}^{\text{cr}}_c = \beta\,\mathbf{r}_c + (1-\beta)\,\mathbf{t}_c,
\end{equation}
and retrieve a recall-oriented candidate set $\mathcal{R}^{\text{cr}}_c$.

\subsubsection{Candidate pool construction}
To construct the final retrieval set $\mathcal{R}_c$ for a given category $c$, we perform a score-maximizing union of the subsets retrieved via both pathways. To resolve collision mappings, we apply a deterministic deduplication that retains the maximum similarity score for each unique asset $x_i \in \mathcal{R}_c$, followed by a top-$K$ truncation to bound downstream computational complexity. Retrieved candidates are filtered and combinatorially reranked by an agentic VLM. This formulation yields an over-inclusive, high-recall buffer.

\subsection{Avatar Assembly and Execution}
\label{sec:execution_vqa}
Given the candidate pool ${\mathcal{R}}_c$ and the prompt-conditioned concept scaffold (Sec.~\ref{sec:concept_scaffold}), we assemble executable 3D avatars through a VLM-driven selection and verification pipeline that generates multiple candidate looks per prompt and selects the best via batched VLM comparison.

\subsubsection{Cohesive assembly from category pools and concept art}
Each category ${\mathcal{R}}_c$ is displayed as a grid of asset thumbnails (one grid per category). These grids are pre-filtered by a VLM to exclude low-quality or meme-like assets (e.g., text-dominated or non-wearable thumbnails), so that the selection stage sees only visually meaningful options. The filtered grids are then presented to a VLM, together with the rendered concept art and the user prompt. The VLM selects at most one asset per category to form a single cohesive outfit, prioritizing the text description and using the concept art as a visual prior for style and layout to guide the selection. This yields an initial asset combination, which is then composed into a 3D look and rendered for verification.

\subsubsection{Iterative VQA and refinement}
Because 2D thumbnail evaluation cannot guarantee 3D topological success, we run an iterative Verification and Quality Assessment (VQA) loop. The composed avatar is rendered and evaluated by the VLM against both the text prompt and the concept art for prompt alignment, completeness, and geometric consistency (e.g., mesh clipping). When the result is insufficient, the VLM proposes concrete edits: add a missing asset, remove an ill-fitting one, or substitute an asset from the same grid. We apply these edits, recompose the look, and re-evaluate until the assessment is satisfactory or a preset maximum iteration count (chosen empirically) is reached.

\subsubsection{Multi-candidate generation and batch selection}
To explore the combinatorial space while avoiding repetitive outputs, we generate multiple candidate looks per prompt. A per-asset and per-bundle usage cap limits how often any single asset or body bundle is reused across candidates, and we rotate over the top few body bundles so that the same refined asset set is composed with different base bundles. This yields a set of candidate looks. We then run a batched VLM comparison over all candidates: in each batch, the VLM is shown multiple avatars at once and selects the best according to visual quality and fidelity to the prompt; the winners advance to the next round, and the process repeats until a single best look is selected as the final avatar for that prompt.

\begin{figure*}[t]
    \centering
    \includegraphics[width=\textwidth]{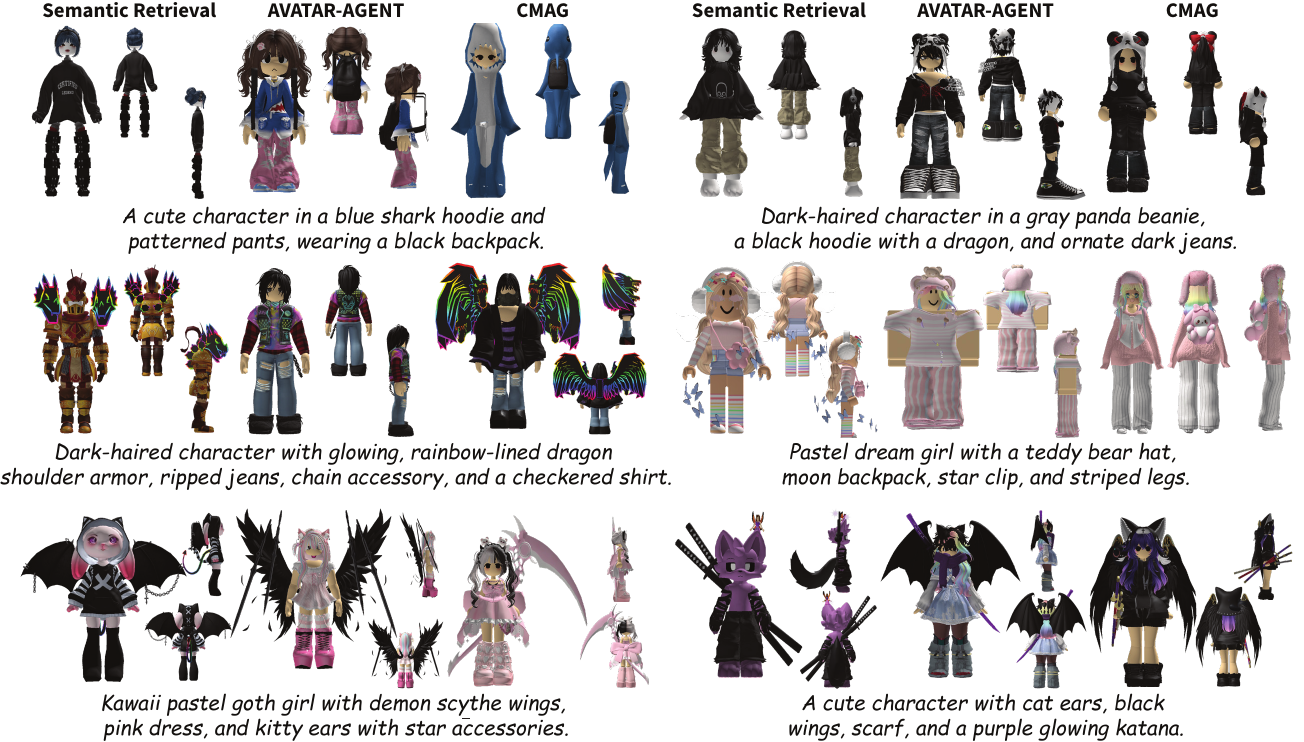}
    \caption{Qualitative comparison between proposed CMAG and state-of-the-art avatar generation methods.}
    \label{fig:sota_comp}
\end{figure*}

\section{Experiments}
\label{sec:experiments}

We implement our framework in PyTorch and fine-tune the Trellis \cite{xiang2025trellis} text-to-3D model using LoRA \cite{hu2022lora} to generate the intermediate 3D concept scaffold. We render four orthogonal views of the scaffold and encode each view with pretrained CLIP (ViT-B/16) image encoder to obtain multi-view global embeddings. We use the CLIP text encoder to embed the routed category-wise text queries, and we encode view-aware segmented part crops with the same CLIP image encoder to obtain localized part embeddings. For text-grounded segmentation, we use SAM 3 \cite{carion2025sam3segmentconcepts}. All agentic components in our pipeline (prompt decomposition, taxonomy routing, candidate filtering, and VQA-based refinement) are implemented using the GPT-5-mini~\cite{singh2025openai} VLM. To mitigate part discovery failures, we apply the feature suppression (concept-residual) strategy following \cite{dorfman2025ip}. We precompute CLIP embeddings for all curated high-quality assets in the corpus and build category-wise FAISS indices \cite{douze2025faiss}; we additionally precompute the category projection matrices used by the feature suppression module. Unless otherwise stated, we set the fusion weights to $\alpha=\beta=0.7$. During retrieval, we form a candidate pool by combining results from both retrieval branches, de-duplicating by retaining the maximum similarity score per asset, and keeping the top-$40$ candidates per category. The visual-linguistic gating agent then retains the top-$20$ semantically correct and stylistically compatible assets per category for downstream avatar assembly.

\subsection{Comparison with State-of-the-Art Methods}
\label{sec:sota_comp}

We evaluate CMAG against two baselines: (i) AVATAR-AGENT~\cite{ding2026avataragent}, a state-of-the-art multi-agent framework for expressive 3D avatar generation, and (ii) a semantic retrieval baseline that performs CLIP-based similarity search over the production catalog using a linear combination of text and image embeddings. All methods are evaluated on a curated suite of 200 complex compositional prompts designed to stress-test semantic ambiguity and stylistic diversity. Retrieval is performed over a large-scale production catalog containing 100,000 discrete, creator-generated 3D assets. Each prompt requires the system to select and compose multiple compatible marketplace assets from this extensive catalog under strict platform taxonomy constraints.

\noindent\textbf{Quantitative Evaluation.}
We report standard evaluation metrics, including CLIPScore~\cite{radford2021learning}, Aesthetics Score~\cite{duggal2025eval3d}, and Text-to-3D alignment~\cite{duggal2025eval3d}. For Text-to-3D alignment, we use GPT-5.2 as an external evaluator to decouple generation from evaluation and reduce self-evaluation bias. A fixed evaluation rubric and prompt template are used across all methods.

\noindent\textbf{Results.}
Table~\ref{tab:sota} summarizes our quantitative findings. CMAG achieves the strongest overall performance, attaining a CLIPScore of 30.7, a Text-to-3D alignment of 47.0, and an Aesthetics score of 82.0. Compared to AVATAR-AGENT, CMAG maintains comparable visual quality while improving CLIPScore from 30.4 to 30.7 and increasing Text-to-3D alignment from 43.0 to 47.0. The performance gap is larger relative to the semantic retrieval baseline, where CMAG improves Text-to-3D alignment from 33.9 to 47.0 and Aesthetics from 79.0 to 82.0, alongside a gain in CLIPScore from 30.2 to 30.7.

These improvements in Text-to-3D alignment indicate that CMAG more reliably satisfies complex prompt attributes and compositional constraints. The strong Aesthetics score further suggests that concept-scaffolded retrieval and taxonomy-aware routing produce visually coherent and stylistically consistent avatar assemblies. Overall, the results support our hypothesis that incorporating 3D concept scaffolds provides critical visual context, reduces semantic ambiguity, and enhances retrieval robustness for marketplace-constrained generation.

\begin{table}[t]
  \centering
  \caption{Comparison of CLIP Score, Text-to-3D alignment and Aesthetics Score. Best results are shown in \textbf{bold}.}
  \label{tab:sota}
  \resizebox{\linewidth}{!}{
  \begin{tabular}{lccc}
    \toprule
    Method & CLIP Score $\uparrow$ & Text-to-3D Align. $\uparrow$ & Aesthetics $\uparrow$ \\
    \midrule
    Semantic Retrieval          & 30.2 & 33.9 & \underline{79.0} \\
    AVATAR-AGENT & \underline{30.4} & \underline{43.0} & \textbf{82.0} \\
    \midrule
    CMAG (Ours)  & \textbf{30.7} & \textbf{47.0} & \textbf{82.0} \\
    \bottomrule
  \end{tabular}
  }
  \vspace{-1.2em}
\end{table}

\subsection{Ablation Study}
\label{sec:ablation}

We conduct ablation studies to validate three key design choices in CMAG: (i) concept scaffolding for ambiguity-resilient retrieval, (ii) taxonomy-aware routing for category coverage and topological coherence, and (iii) feature suppression for robust body bundle retrieval. As summarized in Table~\ref{tab:ablation}, removing any of these components consistently degrades performance across all metrics, with the most significant drops observed in Text-to-3D alignment.

\begin{table}[h]
\centering
\caption{Comparison of CLIP Score, Text-to-3D alignment and Aesthetics Score across method variants.}
\label{tab:ablation}
\resizebox{\columnwidth}{!}{%
\begin{tabular}{lccc}
\toprule
\textbf{Method Variant} & \textbf{CLIPScore $\uparrow$} & \textbf{Text-to-3D Align. $\uparrow$} & \textbf{Aesthetics $\uparrow$} \\
\midrule
w/o Taxonomy Policy Agent & 30.0 & 40.0 & 80.0 \\
w/o 3D Concept Scaffold & 30.2 & 41.0 & 81.0 \\
w/o Feature Suppression & 30.5 & 43.0 & 82.0 \\
\midrule
\textbf{CMAG (Full)} & \textbf{30.7} & \textbf{47.0} & \textbf{82.0} \\
\bottomrule
\end{tabular}%
}
\vspace{-1em}
\end{table}

\subsubsection{Effect of the 3D concept scaffold}
\label{sec:ablation_scaffold}

\begin{figure}[t]
    \centering
    \includegraphics[width=\columnwidth]{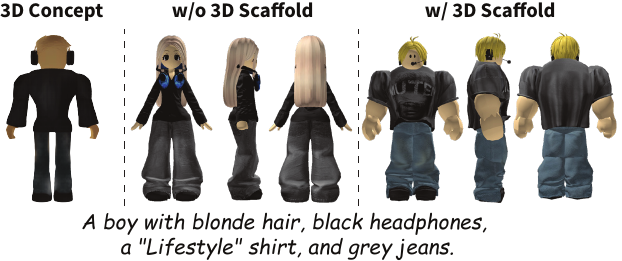}
    \caption{
    \textbf{Effect of the 3D concept scaffold on retrieval and composition.}
    Given the prompt, text-only retrieval (middle) drifts to semantically incorrect assets, missing key attributes and selecting incompatible components. Incorporating the 3D concept scaffold (right) provides multi-view visual priors that guide retrieval toward a prompt-faithful and coherent avatar.
    }
    \label{fig:concept_scaffold}
    \vspace{-0.5em}
\end{figure}

To evaluate the contribution of the 3D concept scaffold beyond language supervision, we disable prompt-to-3D generation and multi-view encoding, and instead perform text-only retrieval using routed text queries and CLIP text embeddings while keeping the remaining pipeline unchanged. As shown in Fig.~\ref{fig:concept_scaffold}, removing the scaffold leads to retrieval drift under prompt ambiguity: the system selects visually plausible but semantically incorrect assets and misses key attributes specified in the prompt. In the illustrated example, the text-only variant retrieves an avatar with blue headphones and a black jacket, failing to match the requested blonde hair, black headphones, ``Lifestyle'' shirt, and grey jeans. Quantitatively, this loss of visual context causes a significant drop in Text-to-3D alignment (from 47.0 to 41.0) and CLIPScore (from 30.7 to 30.2). In contrast, incorporating the 3D concept scaffold provides a multi-view visual prior over silhouette, garment layout, and accessory placement. This additional visual context disambiguates intent and guides retrieval toward assets that more faithfully match the prompt, resulting in a coherent avatar with the correct hair color, accessories, and clothing details.

\subsubsection{Effect of taxonomy policy agent}
\label{sec:ablation_taxonomy}

\begin{figure}[t]
    \centering
    \includegraphics[width=\columnwidth]{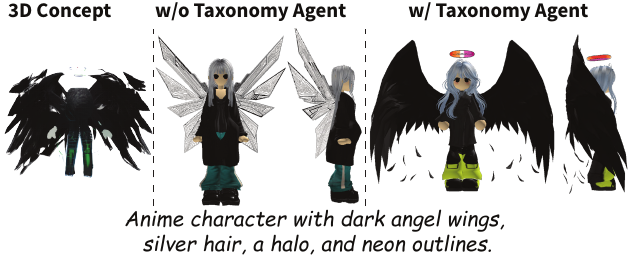}
    \caption{
    \textbf{Effect of the taxonomy policy agent.}
    Given the prompt, removing taxonomy routing (middle) results in retrieval of an incorrect wing category, omission of the halo accessory, and missing neon stylistic elements. Enabling the taxonomy agent (right) resolves semantic-to-taxonomic ambiguity, enforces category coverage, and yields a prompt-faithful composition.
    }
    \label{fig:tax_agent_qual}
\vspace{-1em}
\end{figure}

We evaluate the impact of explicit taxonomy routing by replacing the taxonomy policy agent with a naive category selection strategy (e.g., selecting a single best-guess category per concept) while keeping retrieval and verification fixed. As shown in Fig.~\ref{fig:tax_agent_qual}, removing the taxonomy agent leads to semantic-to-taxonomic mismatch and incomplete category coverage. In the illustrated example, the system retrieves an incompatible wing category, omits the halo accessory, and fails to reflect the requested neon highlights, despite these attributes being explicitly specified in the prompt. This structural and stylistic failure is reflected quantitatively by a severe drop in Text-to-3D alignment from 47.0 to 40.0, alongside decreases in both CLIPScore (30.0) and Aesthetics (80.0). With taxonomy routing enabled, ambiguous concepts are expanded into the appropriate category set and mutually exclusive decisions are enforced prior to retrieval. This ensures that all required categories (e.g., wings, halo, stylistic effects) are explicitly considered, producing a composition that faithfully satisfies both structural and stylistic constraints of the prompt.

\subsubsection{Effect of feature suppression (concept-residual) on body retrieval}
\label{sec:ablation_suppression}

\begin{figure}[t]
    \centering
    \includegraphics[width=\columnwidth]{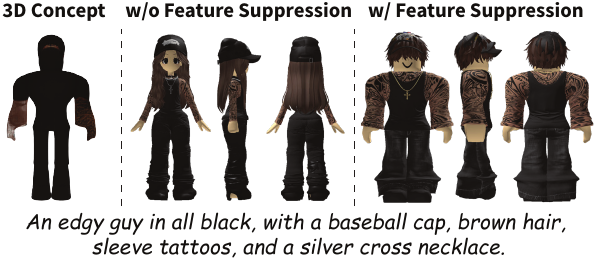}
    \caption{
    \textbf{Effect of feature suppression on body retrieval.}
    Given the prompt, retrieval without feature suppression (middle) selects a body bundle that is inconsistent with the intended identity despite correct clothing and accessories. Incorporating feature suppression (right) removes interfering category directions in the global embedding, leading to a body bundle that better matches the scaffold and prompt constraints.
    }
    \label{fig:feature_suppression}
\vspace{-1em}
\end{figure}

We analyze the role of feature suppression for body bundle retrieval, especially when localized part evidence is missing or unreliable. To ablate this module, we query the body category using the fused global embedding without orthogonalizing against non-target category subspaces. As shown in Fig.~\ref{fig:feature_suppression}, without suppression the global representation is often dominated by salient clothing and accessory cues, which bias retrieval toward body bundles inconsistent with the scaffold’s silhouette and intended identity attributes. In the illustrated example, although clothing items are correctly retrieved, the body bundle does not align with the prompt-specified identity. Without this module, the Text-to-3D alignment drops from 47.0 to 43.0 and the CLIPScore decreases to 30.5. By suppressing non-target category directions, feature suppression isolates body-specific cues within the global embedding. This leads to retrieval of a body bundle that better matches the scaffold’s structure and the prompt’s identity constraints, producing a more coherent final avatar.

\section{Conclusion and Future Work}
\label{sec:conclusion}

We introduced CMAG, a 3D-aware framework for composing high-quality, topologically consistent avatars from discrete marketplace assets using complex prompts. CMAG leverages an intermediate 3D concept scaffold to disambiguate text, performs taxonomy-aware routing to ensure category coverage, and combines hybrid retrieval with vision--language verification to assemble prompt-faithful, executable avatars under platform constraints. While effective, our current CLIP-based embeddings are fundamentally 2D. Projecting 3D geometry into limited rendered views discards depth, articulation, and structural attachment semantics. Moreover, CLIP's sensitivity to rendering factors (e.g., scale, pose, background) introduces noise that may degrade retrieval for small or occluded assets. A promising future direction is developing 3D-native, viewpoint-robust representations that explicitly encode avatar compositional structure and attachment semantics, such as distinguishing waist from back accessories or predicting geometric conflicts between items. Such structured representations could further improve retrieval robustness, enable interpretable selection, and strengthen creator-driven metaverse ecosystems by making UGC assets reliably discoverable and composable.

{
    \small
    \bibliographystyle{ieeenat_fullname}
    \bibliography{main}
}

\end{document}